\documentclass[letterpaper, 10 pt, conference]{ieeeconf} % Comment this line out if you need a4paper

\IEEEoverridecommandlockouts % This command is only needed if 
\usepackage{cite}
\usepackage{graphicx} % for pdf, bitmapped graphics files
\graphicspath{{./figures/}}
\usepackage{epsfig} % for postscript graphics files
\usepackage{mathptmx} % assumes new font selection scheme installed
\usepackage{times} % assumes new font selection scheme installed
\usepackage{amsmath} % assumes amsmath package installed
\usepackage{amssymb} % assumes amsmath package installed
\usepackage[skip=0.3ex]{subcaption}
\usepackage{placeins}
\usepackage{float}
\usepackage[font=small]{caption}
\usepackage{subcaption}
\usepackage{xcolor}
\usepackage{dsfont}
\usepackage{amsfonts}
\usepackage{bm}
\usepackage{enumerate}
\usepackage{subeqnarray}
\usepackage{algorithmic}
\usepackage{algorithm}
\usepackage{indentfirst} 
\usepackage{cite}
\usepackage{tikz}
\usepackage{multirow}
\usepackage{mathtools}
\usepackage{siunitx}
\usepackage{url}
\usepackage[outdir=./figures]{epstopdf}
\newtheorem{definition}{Definition}

\newtheorem{althm}{Algorithm}
 \DeclareMathAlphabet      {\mbi}{OML}{cmm}{b}{it}
\DeclareMathAlphabet      {\mathit}{OML}{cmm}{}{it}
\newcommand{\oB}{\begin{bmatrix}}
\newcommand{\cB}{\end{bmatrix}}

\DeclareMathAlphabet      {\mathit}{OML}{cmm}{}{it}
\DeclareMathAlphabet{\mathcal}{OMS}{cmsy}{b}{n}

\newcommand{\R}{\mathbb{R}}
\newcommand{\Q}{\mathcal{Q}}
\newcommand{\TQ}{\text{T}\mathcal{Q}}

\begin{document}

\title{\LARGE \bf EigenMPC: An Eigenmanifold-Inspired Model-Predictive Control Framework for Exciting Efficient Oscillations in Mechanical Systems}

%\author{Andre Coelho${}^{1,2}$%
\author{Andre Coelho${}^{1,2,3}$, Alin Albu-Schaeffer${}^{1,4}$, Arne Sachtler${}^{1,4}$, Hrishik Mishra${}^{1,5}$, Davide Bicego${}^3$, \\ Christian Ott${}^{1,5}$, Antonio Franchi${}^{3,6}$% <-this % stops a space
\thanks{${}^1$Institute of Robotics and Mechatronics of the German Aerospace Center (DLR), Oberpfaffenhofen, Germany.}%
\
\thanks{${}^2$Dextrous Robotics Inc., Memphis, TN, USA.}%
\
\thanks{${}^3$Robotics and Mechatronics Lab, Faculty of Electrical Engineering, Mathematics \& Computer Science, University of Twente, Netherlands.}%
\
\thanks{${}^4$Department of Informatics, Technical University of Munich, Germany.}%
\
\thanks{${}^5$Automation and Control Institute (ACIN), Technical University of Vienna, Austria.}%
\
\thanks{${}^6$LAAS-CNRS, University of Toulouse, France.}%
\
\thanks{{\tt\small andre@dextrousrobotics.com}}}%}

\maketitle

\begin{abstract}
This paper proposes a Nonlinear Model-Predictive Control (NMPC) method capable of finding and converging to energy-efficient regular oscillations, which require no control action to be sustained. The approach builds up on the recently developed Eigenmanifold theory, which defines the sets of line-shaped oscillations of a robot as an invariant two-dimensional submanifold of its state space. By defining the control problem as a nonlinear program (NLP), the controller is able to deal with constraints in the state and control variables and be energy-efficient not only in its final trajectory but also during the convergence phase. An initial implementation of this approach is proposed, analyzed, and tested in simulation.
\end{abstract}
\FloatBarrier

\IEEEpeerreviewmaketitle

\setlength{\textfloatsep}{10pt}

\section{Introduction}
In the last three decades, numerous roboticists have devoted their effort to generating energy-efficient robot motion \cite{mcgeer1990passive, bjelonic22}. The developed approaches are especially useful to render cyclic motions, like walking gaits and pick-and-place trajectories. In most cases, mechanisms and control laws are designed to make use of the shape of the gravitational or elastic potential in order to achieve maneuvers requiring little to no control effort to be sustained. 

\begin{figure}[t]
\centering
 \includegraphics[trim={1.8cm 2cm 1.5cm 5cm},clip,width=0.88\linewidth]{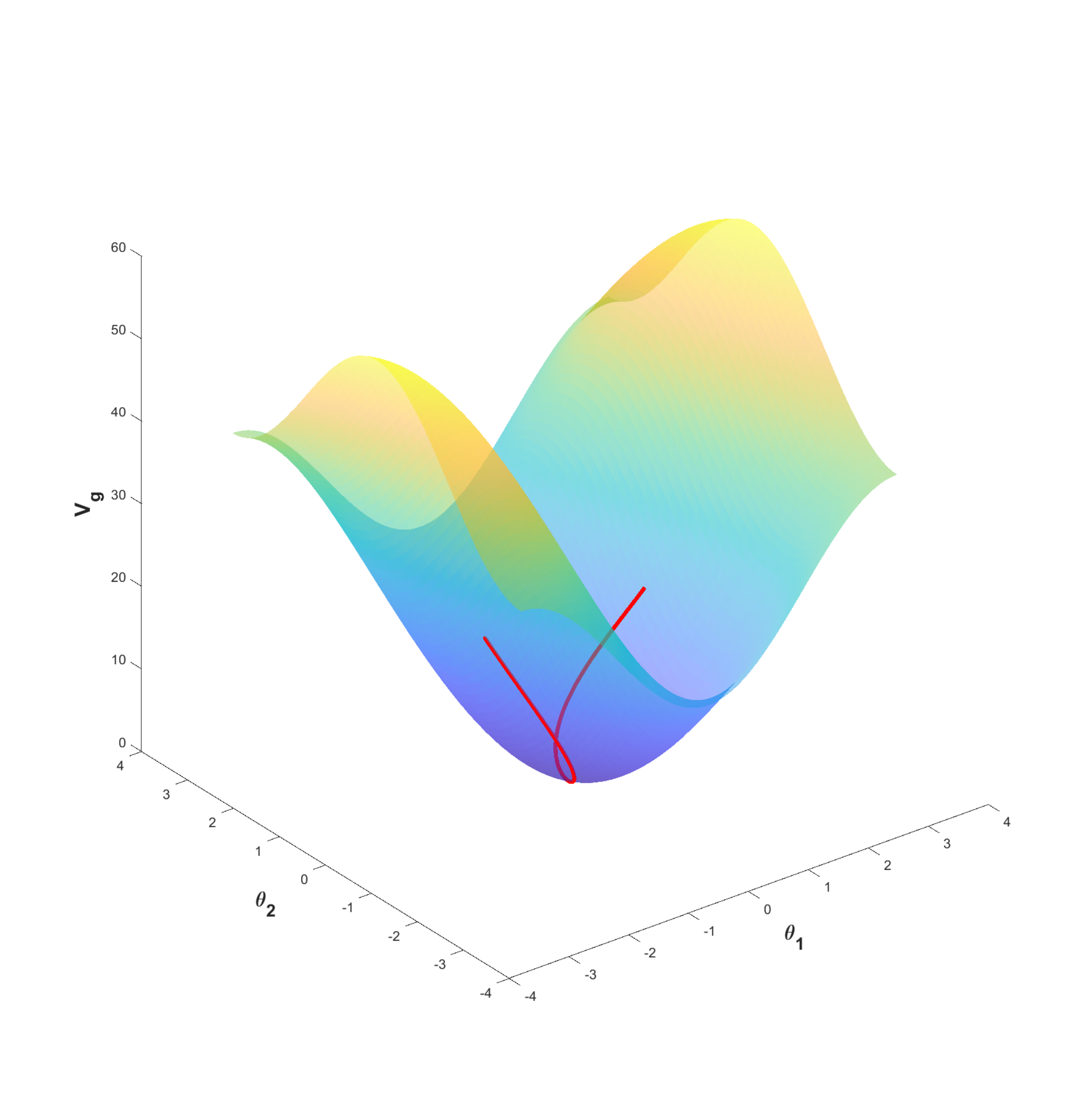}
\caption{An eigenmode (red line) on the potential energy surface.}
\label{fig:surface}
\end{figure}

In that scope, recent works have extended the concept of normal modes to nonlinearly-coupled mechanical systems \cite{kerschen2009nonlinear, albu-schaeffer2020review}. In particular, \cite{albu-schaeffer2020review} has demonstrated a way of finding sets of invariant line-shaped oscillations, called eigenmodes (see Fig.~\ref{fig:surface}), which belong to a two-dimensional submanifold of the state space: the eigenmanifold. \textcolor{black}{The two properties of eigenmodes, namely invariance and line shape, are of paramount importance for robotic tasks like pick-and-place and locomotion. The former guarantees that no energy injection is needed to sustain a gait, while the latter ensures the oscillations happen between two static poses, which could be for instance the grasping and releasing poses or the leg-lift and ground-contact ones.}

Once such regular oscillation modes are found, proper controllers should be designed in order to drive mechanical systems from their initial state onto the eigenmanifold. In \cite{della-santina21}, a chart to the eigenmanifold is defined and the controller acts along the directions normal to a desired eigenmode to achieve convergence. Although the aforementioned seminal works showed that energy injection is not required to remain on the eigenmanifold, the bounds of the control action needed to converge to them has thus far not been considered. Therefore, such eigenmanifold-stabilization techniques might fail to drive mechanical systems with limited actuation to eigenmodes of arbitrary energy. An example thereof is the DLR Suspended Aerial Manipulator (SAM) \cite{sarkisov19}, \textcolor{black}{which, for safety and energy-efficiency reasons, hangs from a carrier in a pendulum-like setup}. There, the mounted propellers are able to stabilize the system around its hanging point (see \cite{sarkisov20icra}), but not to bring it to arbitrary poses in a static way. On the other hand, proper control design may allow the SAM to reach high-amplitude oscillations and subsequently perform dynamic grasping or perching maneuvers despite its actuation limitations.
\par
In contrast to state-of-the-art eigenmanifold-stabilization methods, Nonlinear Model Predictive Control (NMPC) \cite{gruene17} -- as an optimization-based framework -- is able to deal with constraints of different sorts in the control signal, including propeller-thrust limitations (see \cite{bicego20nonlinear}). In addition, some versions of NMPC (e.\,g., Economic NMPC \cite[Sec.~7.10]{gruene17}) have also achieved periodic orbit stabilization. However, there is no guarantee that the achieved orbits will present the aforementioned properties of eigenmodes, namely invariance (or zero control) and line shape.
\par
In light of that, we propose a novel NMPC framework, which, endowed with knowledge from the eigenmanifold theory, is able to converge to energy-efficient line-shaped oscillatory trajectories on the eigenmanifolds. \textcolor{black}{The proposed approach can deal with actuation bounds while still being able to converge to regular oscillations in an close-to-optimal manner. Moreover, a heuristic version of the framework which does not require previous computation of the eigenmanifold is also introduced. Rather, an estimate of the direction of the mode -- given by its associated eigenvector -- is sufficient for the online search and convergence process.}
\par
The controller efficacy and performance -- both with and without previous knowledge of the eigenmode -- are demonstrated through its application to the simple but insightful example of the torque-limited double pendulum. Simulation results confirm that the controller is able to drive the system to trajectories that are almost identical to the eigenmodes found with \cite{albu-schaeffer2020review}, even when starting from significantly large initial conditions.

\section{Geometric Mechanics and Eigenmodes}
\label{sec:eigmanifold}
Recent development in the field of oscillatory normal modes has achieved a generalization of linear modes to nonlinear, elastically \cite{kerschen2009nonlinear} and inertially-coupled \cite{albu-schaeffer2020review} mechanical systems. In particular, by finding invariant line-shaped curves in the state-space, called eigenmodes, the framework presented in \cite{albu-schaeffer2020review} is able to realize sustained regular oscillations in otherwise chaotic systems (e.\,g., robots). In that scope, this section aims at presenting important aspects of \cite{albu-schaeffer2020review}, upon which the EigenMPC approach is built.

\subsection{Definitions}

\begin{definition}[Forced simple mechanical system]
A \textbf{forced simple mechanical system} \cite{bullo2019geometric} is a tuple ($\TQ, \mathbb{G},\mathsf{V},\mathsf{F}$), where $\TQ$ is the tangent-bundle manifold containing all possible trajectories of the system, $\mathbb{G}$ is the Riemannian metric induced by the system's inertia, $\mathsf{V}:\Q\rightarrow\R^{+} \cup \{0\}$ is a potential function, and $\mathsf{F}:\R \times \TQ \rightarrow \text{T}^*\Q$ is the function of control and external wrenches, where $\text{T}^*\Q$ is the cotangent bundle. 
\par
The trajectories of such systems, $\gamma(t) \in \Q$, follow the so-called forced geodesic equation:
\begin{equation}
    {\mathop{\nabla}\limits^{\scriptscriptstyle\mathbb{G}}}_{\dot{\gamma}(t)}\dot{\gamma}(t)=-\text{grad}\mathsf{V}(\gamma(t))+\mathbb{G}^\sharp\left(\mathsf{F}\left(t,\gamma(t),\dot{\gamma}(t)\right)\right) \, ,
    \vspace{0mm}
    \label{eq:forced_geodesic}
\end{equation}
where $\mathbb{G}^\sharp$ is the sharp map, which transforms a force into an acceleration according to the $\mathbb{G}$ metric. In local coordinates $(x,\dot{x}) \in \R^n \times \R^n$ of the manifold $\TQ$, \textcolor{black}{locally representing its generalized position and velocity}, Eq. \ref{eq:forced_geodesic} translates into the well-known manipulator equation
\begin{equation}
    \ddot{x}=M(x)^{-1}\left(-C(x,\dot{x})\dot{x}-\frac{\partial{V(x)}}{\partial{x}}+F(t,x,\dot{x})\right) \, ,
    \label{eq:manipulator}
\end{equation}
where $F(t,x,\dot{x})$ are the control and external forces, and ${\partial{V(x)}}/{\partial{x}}$ are forces induced by the potential $\mathsf{V}$, which are usually either elastic and/or gravitational forces. 
\end{definition}
\begin{definition}[Energy]
Given a forced mechanical system with $\mathsf{F} \equiv 0$ we locally define its energy as
\begin{equation}
  E(x,\dot{x})=\dfrac{1}{2}\dot{x}^\top M(x)\dot{x}+V(x) \, ,
  \label{eq:energy}
\end{equation}
such that there exists a stable equilibrium point $(x_{eq},0) \in \R^n \times \R^n$ which is a minimum for $E(x,\dot{x})$. 
% In coordinates, we might use
% \begin{equation}
%   E(x,\dot{x})=\dfrac{1}{2}\dot{x}^\top M(x)\dot{x}+V(x) \, 
%   \label{eq:energy}
% \end{equation}
% as a local representation of \eqref{eq:energy}.
\end{definition}
% \begin{definition}[Forced geodesic spray]
% In coordinates, \eqref{eq:forced_geodesic} becomes a second-order differential equation. Therefore, in the absence of forces $\mathsf{F}\left(t,\gamma(t),\dot{\gamma}(t)\right)$, it is possible to infer the existence of a time-independent vector field $S$ acting on $\text{T}\Q$, whose integral curves $\gamma(t)$ result from the action of the potential field on the geodesic equation \eqref{eq:geodesic}. Thus, we call this vector field the \textbf{forced geodesic spray}.
% \end{definition}

\begin{definition}[Linear modes]
Let $\dot{y}=Ay$ be the linearized version of \eqref{eq:manipulator} around a stable equilibrium $(x,\dot{x})=(x_{eq},0)$, $F=0$ with $y=\begin{bmatrix}x^\top & \dot{x}^\top\end{bmatrix}^\top \in \R^{2n}$. Then, there exists $n$ two-dimensional invariant subspaces $ES \simeq \R^{2n}$, with structure 
\begin{equation}
    ES=\text{span}\left((c,0), \, (0,c)\right) \, ,
    \label{eq:ES}
\end{equation}
where $c \in \mathbb{R}^n$ are unit vectors in the local chart of $\mathcal{Q}$. Such spaces are called eigenspaces. Moreover, oscillations along the directions $c$ are called \textbf{linear modes}.
\end{definition}

\begin{definition}[Forced geodesic flow]
Let $\Phi^S_{t_0,t_f}(q,v_q)$ be the maximal integral curve of a forced simple mechanical system starting at $(q,v_q) \in \TQ$ at time $t_0$ and ending at time $t_f$. The projection $\pi_{\Q} \circ \Phi^S_{t_0,t_f}(q,v_q)$ of the integral curve onto $\Q$ is called the \textbf{forced geodesic flow}. In other words, the forced geodesic flow is the path in $q$ the system travels starting at $(q,v_q)$ at $t=t_0$ and ending at $t=t_f$ when $\mathsf{F}\equiv0$.
\end{definition}

\begin{definition}[Eigenmodes and eigenmanifold]
\label{def:eigenmodes}
Let
\begin{equation}
    \gamma_{\mathfrak{M}}(q_0,t)\triangleq \Phi^S_{0,t}(q_0,0)\, , \,\, q_0 \in \Q, \, t \in \R  \,, 
\end{equation}
be curves in $\TQ$ such that:
\begin{enumerate}[(i)]
    \item $\exists T \in \mathbb{R}^{+} \text{, s.t } \gamma_{\mathfrak{M}}(q_0,t)=\gamma_{\mathfrak{M}}(q_0,t+kT),\,\,\forall k \in \mathbb{Z}$, and
    \item the map $O:\pi_{\Q}\circ\gamma_{\mathfrak{M}}(q_0,t) \rightarrow [0,1]$ exists and is a diffeomorphism.
\end{enumerate}
Then, $\gamma_{\mathfrak{M}}$ are called the \textbf{eigenmodes} of ($\TQ, \mathbb{G},\mathsf{V},\mathsf{F}$) and the manifold that collects all $\gamma_{\mathfrak{M}}$ from the same family is called the \textbf{eigenmanifold}, denoted $\mathfrak{M}$.
\par
In other words, $\mathfrak{M}$ is an invariant two-dimensional manifold corresponding to the line-shaped periodic trajectories (eigenmodes) of ($\TQ, \mathbb{G},\mathsf{V},\mathsf{F}$) when $\mathsf{F}\equiv0$. 

Finally, as shown in \cite{albu-schaeffer2020review}, it is possible to find local charts $X:\R^2 \to \R^n$ and $\dot{X}:\R^2 \to \R^n$ with coordinates $(x_m,\dot{x}_m) \in \R^2$ such that $X(x_m,\dot{x}_m)=x$ and $\dot{X}(x_m,\dot{x}_m)=\dot{x}$ for $(x,\dot{x}) \in \mathfrak{M}$. Moreover, for linear systems, $(x_m, \dot{x}_m)=(c^\top x, c^\top \dot{x})$, and $(c^\top X,c^\top \dot{X})$ is the identity function.
\end{definition}

\subsection{Energy as similarity function}
Due to the periodicity of  $\gamma_{\mathfrak{M}}(q_0,t)$, we may define an equivalence relation to drop its dependency on $t$, calling it $\gamma_{\mathfrak{M}}(q_0)$, i.e., disregarding the number of periods of a trajectory and focusing only on its image in the state space as a set. It is also easy to conclude that due to the absence of energy injecting or dissipating forces $\mathsf{F}$, every point in $\gamma_{\mathfrak{M}}(q_0)$ belongs to the same level set of the energy function \eqref{eq:energy}, which is uniquely defined by the energy of the point $(q_0,0) \in \TQ$.

\textcolor{black}{In addition, \cite{albu-schaeffer2020review} defines a relationship of continuity between points $(q_0,0) \in \TQ$ including $(q_{eq},0)$ whose forced geodesic flow generates the eigenmodes, called the generator set $\mathfrak{R}$. This means that, in an arbitrarily small neighborhood of every point in $\mathfrak{R}$, there are other points with different energy levels, which also belong to $\mathfrak{R}$ and, consequently, to an eigenmode of the mechanical system. Through this continuity relationship, the closer a point  $(q_0,0) \in \mathfrak{R}$ is to the equilibrium $(q_{eq},0)$, the more the eigenmode associated with it will resemble one of the straight modes of the linearized version of the system. Thus, we can define the energy of an eigenmode $\gamma_{\mathfrak{M}}(q_0)$ as its degree of similarity with an eigenvector. This property will be used in Section~\ref{sec:straight} to endow the EigenMPC framework with online eigenmode-search capabilities. The idea is to penalize the distance to the eigenvector for low energies and gradually relax the penalization as energy increases.}

\textcolor{black}{\section{Nonlinear Model Predictive Control}}

Nonlinear Model-Predictive Control (NMPC) is an optimization-based controller which acts as described below (see \cite[Algorithm~3.1]{gruene17}).

\begin{althm}[NMPC]
At each sample $t_n \in [t_0,..,t_{N-1}]$, where $N$ is the length of the prediction horizon, follow the steps:
\begin{enumerate}
    \item Measure the state $z_n \triangleq z(t_n) \in \mathcal{Z}$.
    \item Set $z_0\coloneqq z_n$ and solve
    \begin{subequations}\label{NLP}
\begin{align}
\label{eq:JN}
\min_{u_k} &\:\,\,\,J_N \triangleq \sum_{k=0}^{N-1} \ell(z_u(t_k,z_0),u(t_k))\\
\text{s.t.}\,
&0=z_{k+1}-\phi_k(z_k,u_k),\quad  k=0,1,\ldots,N-1,\label{conti const}\\
&\underline{r}_k\leq r_k(z_k,u_k)\leq \overline{r}_k, \quad \;\; k=0,1,\ldots,N-1,\label{path constraint},
\end{align}
\end{subequations}
where $u_k\triangleq u(t_k)$ is the control variable and $z_k \triangleq z_u(t_k,z_0)$ is the open-loop predicted state when $u_k$ is applied. Moreover, $\ell$ and $J_N$ are called the running and the finite-horizon cost functions, respectively, $\phi_k(z_k,u_k)$ is a one-step state-transition function and \eqref{path constraint} defines lower and upper bounds to a function $r(z_k,u_k)$ of the state and control variables.
\item Denote $u^*$ as the optimal solution and use $u^*(t_0)$ as the control action for the next sampling period.
\end{enumerate}
\end{althm}

\section{The EigenMPC framework}
Based on the eigenmanifold theory and NMPC, we propose a framework that allows constrained mechanical systems to be energy efficient in both converging to and sustaining regular oscillations with different levels of \textit{a priori} knowledge. We call this framework \emph{EigenMPC}: Eigenmanifold-Inspired Model Predictive Control.

The action of EigenMPC is twofold:  (a) converging to a pre-computed eigenmode; (b) finding an approximation of the eigenmode and converging to it. Hereafter, we call (a) the curved version and (b) the \textcolor{black}{straight version}.
\subsection{\textcolor{black}{Curved EigenMPC}}
In case the eigenmodes have been pre-computed (e.g., with \cite{albu-schaeffer2020review}), the curved version of EigenMPC is able to stabilize the system to it in a close-to-optimal manner while taking possible constraints into account. The controller is defined as follows:

\begin{definition}[Curved EigenMPC]
Let the chart defined by $X(x_m)$ and $\dot{X}(x_m,\dot{x}_m)$ yield an approximation of a desired eigenmode in $\mathfrak{M}$, with $c(x_m)$ being the unit-norm tangent vector of $X(x_m)$ at $x_m$. We define the curved version of EigenMPC as the NMPC algorithm with the following running cost:

\begin{align}
\label{eq:cost_curved}
\ell(z, F)&= z^\top W_z z + F^\top W_F F \, ,\\
z&=\begin{bmatrix} E_{ref}-E \\ c_\perp(x_{m})\left(x-X(x_{m})\right)\, \\ c_\perp(x_{m})\left(\dot{x}-\dot{X}(x_{m},\dot{x}_{m})\right) \end{bmatrix}  , 
\label{eq:eig_NLP_curved}
\end{align}
where $c_\perp(x_{m})\left(x-X(x_{m})\right)$ and $c_\perp(x_{m})\left(\dot{x}-\dot{X}(x_{m},\dot{x}_{m})\right)$ are a measure of the distance of $x$ and $\dot{x}$ to the eigenmode along the normal of its tangent vector, given by the multiplication with $c_\perp(x_{m})\triangleq I-c(x_{m})c(x_{m})^\top$. Moreover, the matrices $W_z =\text{diag}(w_E, \, w_x , \, w_x , \, w_{\dot{x}} , \, w_{\dot{x}})
\in \R^{2n+1}$ and $W_F \in \R^n$ are positive definite.
\end{definition}
To show that the proposed method is able to locally drive the system \eqref{eq:manipulator} to the desired eigenmode, we start by noting that \eqref{eq:cost_curved} is a proper cost function for the desired eigenmode, i.e, $\ell=0$ when the system moves on the eigenmanifold and $\ell>0$ otherwise. Moreover, we highlight that the control problem at stake is that of a convergence to a set, where $z=0$, rather than a trajectory-tracking problem.

In sequence, we use the controller defined in \cite{della-santina21} as a baseline and note that its inertia-shaping characteristic removes the need for inertia-weighting in the Lyapunov function. Therefore, for that controller, asymptotic stability can be asserted by using a Lyapunov function with constant diagonal weights like $V(z)=z^\top W_z z$, which is exactly the first term in \eqref{eq:cost_curved}.

Now, we denote $ \ell_{PD}$ and $V_{PD}(z)=z^\top W_z z$ the running cost and its state-related part when the control action from \cite{della-santina21}, denoted $F_{PD}$, is applied. We note that asymptotic stability of $V_{PD}$ implies that for some functions $\alpha_1, \alpha_2 \in \mathcal{K}_\infty$,
\begin{equation}
    \label{eq:stability}
    {\alpha}_1\left(V_{PD}(z_0)\right) \leq V_{PD}(z) \leq {\alpha}_2\left(V_{PD}(z_0)\right) \, ,
\end{equation}
for a connected set of initial conditions $z_0 \in \mathcal{Z}_0 \subseteq \mathcal{Z}$ containing the origin.

In addition, by using the fact that $X$, $\dot{X}$ and $E_{ref}$ parameterize an eigenmode, which by Def.~\ref{def:eigenmodes} requires no control force to be sustained, we get that $z \to 0$ implies $F_{PD} \to 0$, which together with \eqref{eq:stability} implies (local) asymptotic controllability of the system with the \emph{small control property} (see \cite[Def.~4.2]{gruene17}), which states that for each $z \in \mathcal{Z}_0$ there exists $F$ (in this case $F_{PD}$) and $\beta \in \mathcal{KL}$ such that
\begin{equation}
\label{eq:controllability}
    \ell(z,F) \leq \beta\left(V_{PD}(z_0),n)\right) \, .
\end{equation}
Now, we define
\begin{equation}
    \ell^{*}(z)\triangleq \inf_{F \in \R^n} \ell(z,F) \, .
\end{equation}
From the fact that $\ell^{*}(z) \leq \ell(z,F)$, from its quadratic nature and from the small control property \eqref{eq:controllability}, we get
\begin{equation}
    \alpha_3\left(V_{PD}(z)\right) \leq \ell^{*}(z) \leq \alpha_4\left(V_{PD}(z)\right) \, .
\end{equation}
If in addition, there exists a $\beta \in \mathcal{KL}_0$, which is linear in its first argument and summable (see \cite[Ass.~6.4]{gruene17}),  such that, for each $z \in \mathcal{Z}_0$ and for some $F$,
\begin{equation}
\label{eq:beta}
   \ell(z,F) \leq \beta (\ell^{*}(z),n) \, .
\end{equation}
Then, from the principle of relaxed dynamic programming, we use the result from \cite[Th.~6.21]{gruene17} to state that the nominal NMPC closed-loop system with dynamics as \eqref{eq:manipulator} and NMPC-feedback is asymptotically stable on $\mathcal{Z}_0$ provided $N$ is sufficiently large. 

Although hard to verify in practice, the assumption on the linearity of $\beta$ seems reasonable for systems like \eqref{eq:manipulator}, which are feedback linearizable. Further evidence on the convergence capabilities of the approach will be presented in Section~\ref{sec:double-pendulum}.

Note that by using the controller from \cite{della-santina21} as a baseline, the stability analysis is confined to the region of the state-space $\mathcal{Z}_0$ where the that controller is also stable. However, by noting that an infinite-horizon NMPC would find the optimal solution in terms of \eqref{eq:JN} and \eqref{eq:cost_curved} and that the suboptimality number $\alpha \in (0.1]$ of NMPC (see \cite[Th.~4.11]{gruene17}) increases with the size of its horizon, we argue that for sufficiently large $N$, the region of attraction of EigenMPC should be even larger than that of the controller from \cite{della-santina21} and its solution closer to the optimal one.

Moreover, while constraints were not directly accounted for in the analysis, their effect is the following: for bounds on the control action, a connected set $\mathcal{Z}_\mathcal{U} \subseteq \mathcal{Z}_0$ around the eigenmanifold will be attractive, given the zero effort needed to sustain an eigenmode and the continuity of the state space. On the other hand, in case the state is constrained with unconstrained control, a set $\bar{\mathcal{Z}}\subseteq \mathcal{Z}_0$ will be attractive given that the eigenmode itself lies inside the viable set. Finally, in case both state and control are constrained, a set of viable initial states $\bar{\mathcal{Z}}_\mathcal{U}\subseteq \mathcal{Z}_0$ can be attractive. However, there might be cases where the control action is not strong enough to drive the system to an eigenmode without exiting the set of viable states. In this case, the NMPC problem is called inviable (see \cite[Ass. 3.3]{gruene17}).

\begin{figure}[tb]
\centering
 \includegraphics[width=0.9\linewidth]{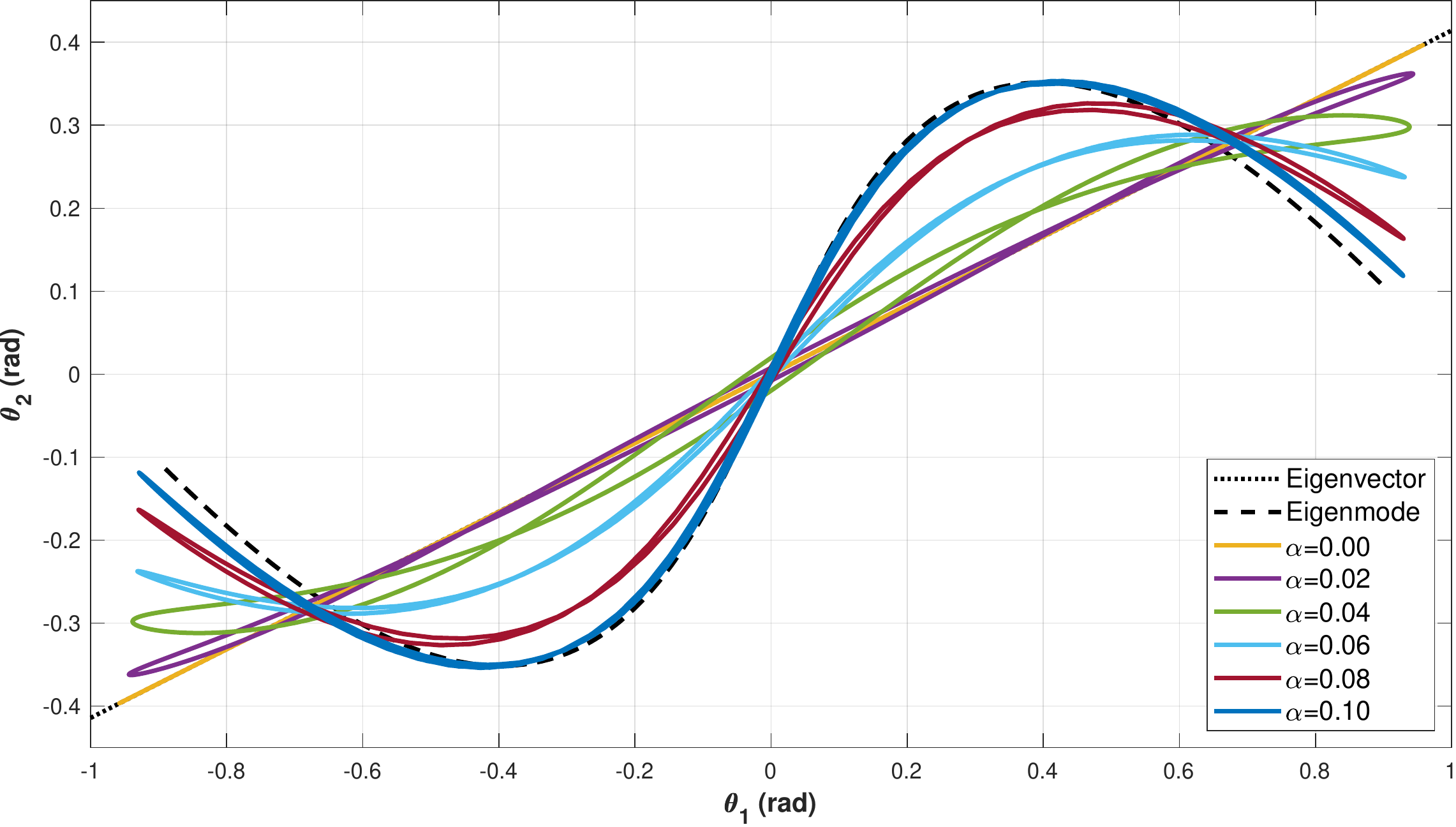}
\caption{Final trajectory of an EigenMPC-controlled double pendulum for different values of $\alpha$. As $\alpha$ increases, the final trajectory moves from a straight line along the linear mode (dotted line) to a curved one near the actual eigenmode (dashed line).}
\label{fig:alpha_varying}
\end{figure}

\subsection{Straight EigenMPC}
\label{sec:straight}
While the curved version of EigenMPC has well-defined stability properties, one might benefit from a more heuristic controller, which does not require the pre-computation of the eigenmanifold. This version, called straight EigenMPC is designed as to require only the direction of its associated eigenvector as \textit{a priori} information.

The online-searching capabilities of EigenMPC are built on the fact that the smaller the energy of an eigenmode of a nonlinear system -- assuming $E(q_{eq},0)=0$ --, the more it resembles an eigenvector of its linearized counterpart. Therefore, we define the straight EigenMPC program as follows:

\begin{definition}[Straight EigenMPC]
Let ES be an eigenspace of a linearized version of \eqref{eq:manipulator} with $c$ as its direction vector (see \eqref{eq:ES}). Then, NMPC with the following running cost is expected to locally drive the system to a neighborhood of the eigenmode associated with ES with a desired energy level $E_{ref}$:
\begin{align}
\label{eq:cost_straight}
\ell(z, F)&= z^\top W_z z + F^\top W_F F \, ,\\
z&=\begin{bmatrix} E_{ref}-E_k \\ c_\perp \left(x-x_{eq}\right)\,(1-\tanh(\alpha E))\, \\ c_\perp \dot{x}\,(1-\tanh(\alpha E))\, \\ F\,(\tanh(\alpha E)+\beta) \end{bmatrix}  , 
\label{eq:eig_NLP_straight}
\end{align}
where $\alpha$ and $\beta$ are scalar tuning factors.
\end{definition}

The chosen cost function can be interpreted in natural language as: ``Reach a desired energy level while finding a trade-off between remaining close enough to the linear mode and finding the paths that minimize the control action.''

Apart from the fixed weights, a gain $\alpha$ is added, which can be understood as an exploration factor. By using $\tanh(\alpha E)$, we ensure a smooth transition between two different behaviors, namely converging to the linear mode and minimizing the control action. The transition occurs as energy increases, at a rate defined by $\alpha$. With higher $\alpha$, the controller gets an exploratory behavior, searching for trajectories with minimum power; whereas with lower $\alpha$, the controller behaves in a conservative way, remaining closer to the linear mode. An example of the controller behavior for different $\alpha$ is shown in Fig.~\ref{fig:alpha_varying}. It can be noted that, with $\alpha=0$ the system converges to the linear mode (dotted line), whereas for increasing $\alpha$ values, the final trajectory approaches the actual eigenmode (dashed line). Moreover, since $\tanh(0)=0$, the factor $\beta$ has been added to the control-torque objective in order to ensure that the control action is not excessively high for low energies.

A significant advantage of the straight EigenMPC over its curved counterpart is that it does not require previous offline computation of the modes. Nevertheless, since the minimum cost $\ell$ is not  exactly on the manifold, but a result of the choice of $\alpha$, we can only guarantee that the controller drives the system to a neighborhood of an eigenmode (see Fig.~\ref{fig:alpha_varying}).

\begin{figure}[tb]
\centering
\begin{subfigure}[t]{0.46\linewidth}
\centering
 \includegraphics[width=0.9\textwidth]{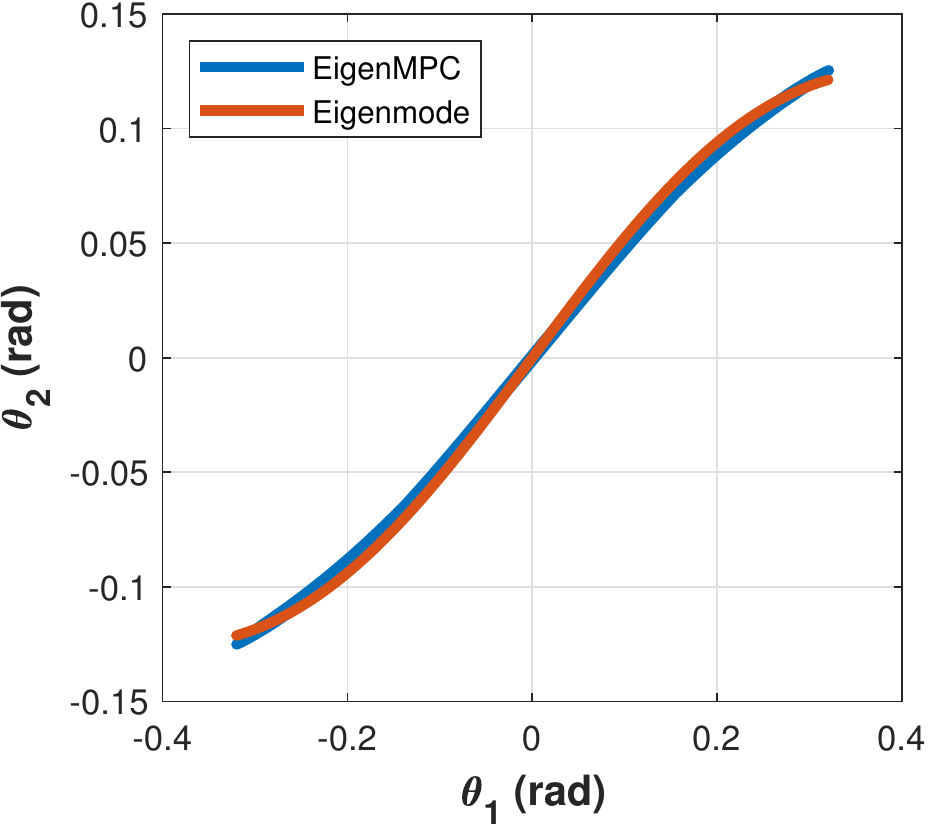}
\caption{In-phase mode, $E_{ref}\approx\SI{2}{J}$.}
\label{subfig:modes_a}
\end{subfigure}%
\begin{subfigure}[t]{0.46\linewidth}
\centering
 \includegraphics[width=0.9\textwidth]{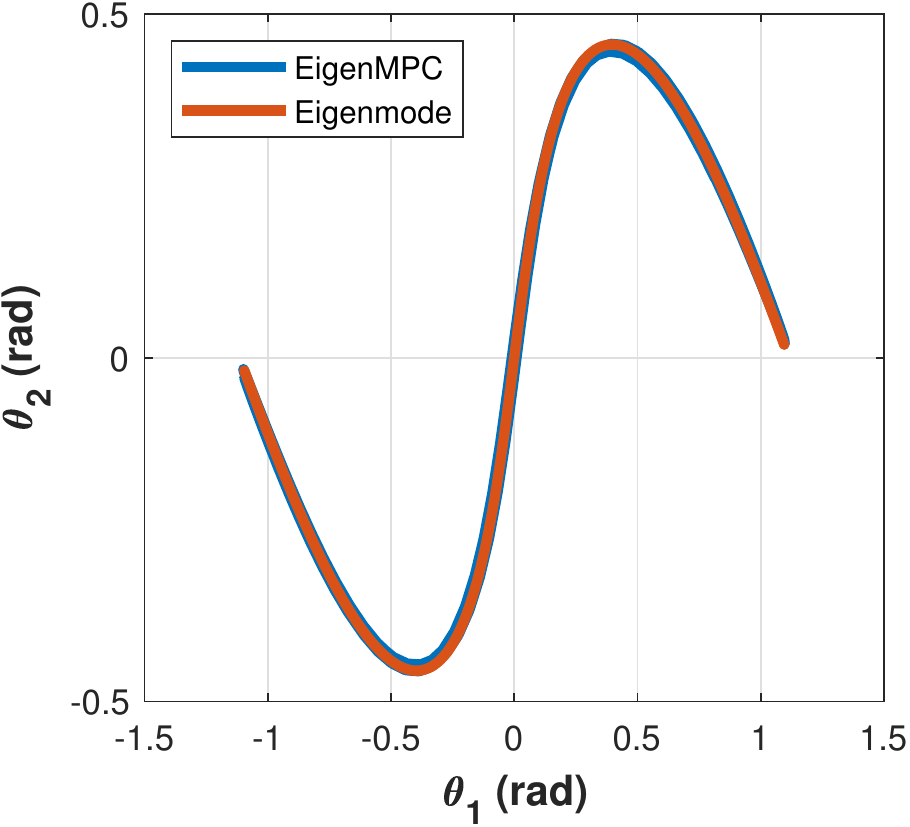}
\caption{In-phase mode, $E_{ref}\approx\SI{16}{J}$.}
\label{subfig:modes_b}
\end{subfigure}\\
\begin{subfigure}[t]{0.46\linewidth}
\centering
 \includegraphics[width=0.9\textwidth]{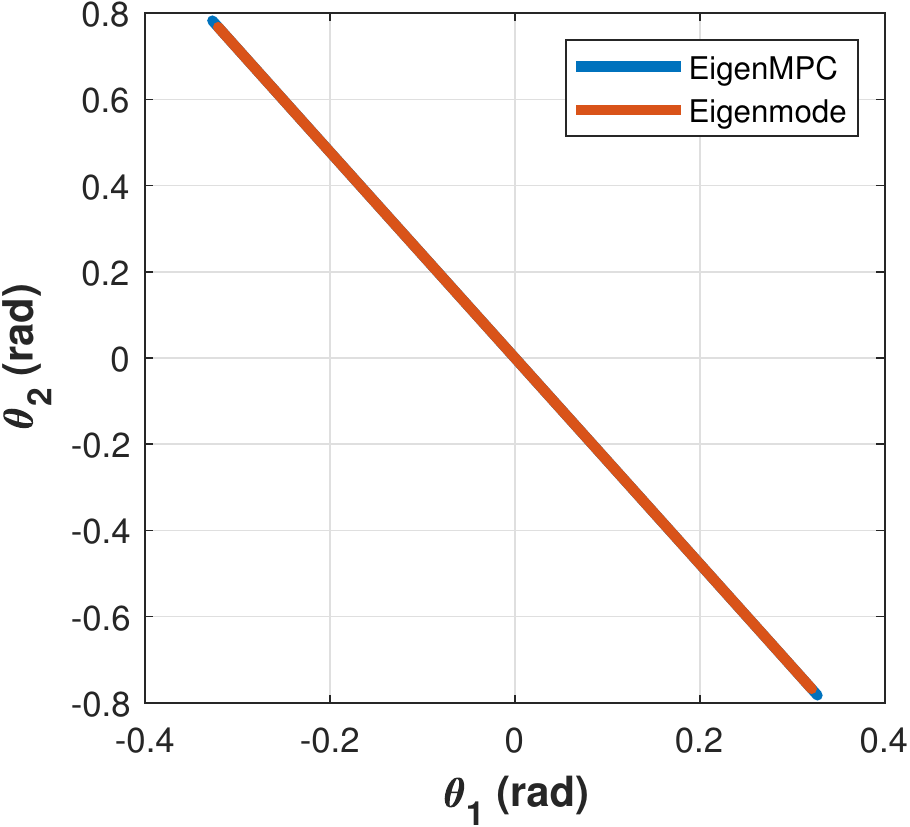}
\caption{Anti-phase mode, $E_{ref}\approx\SI{2}{J}$.}
\end{subfigure}%
\begin{subfigure}[t]{0.46\linewidth}
\centering
 \includegraphics[width=0.9\textwidth]{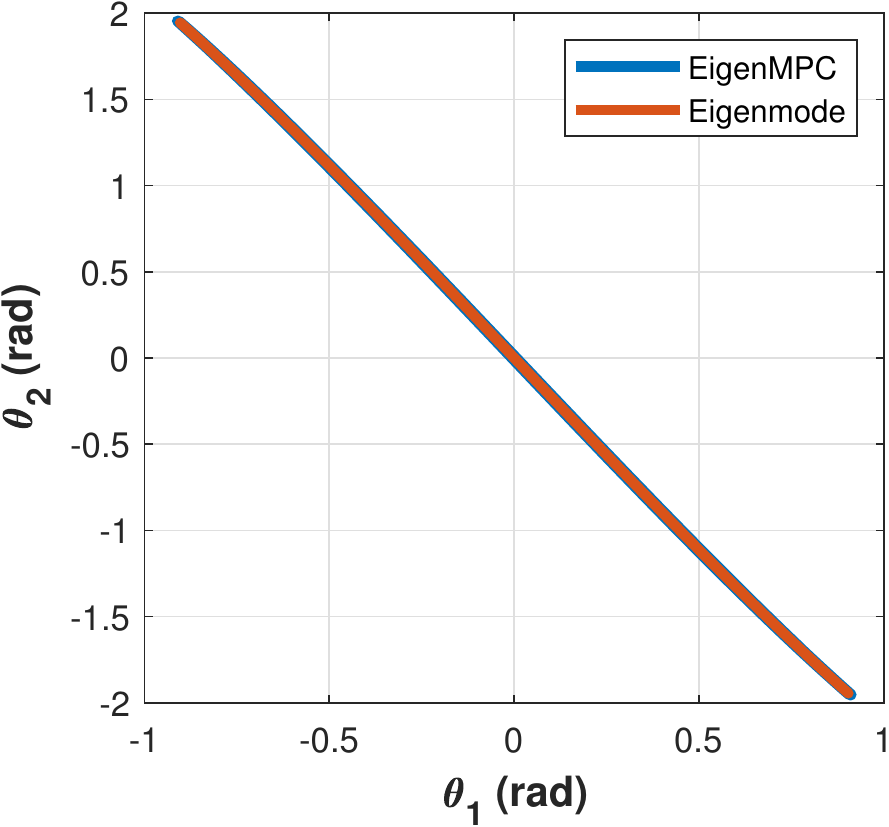}
\caption{Anti-phase mode, $E_{ref}\approx\SI{12}{J}$.}
\end{subfigure}
\caption{Oscillations discovered by EigenMPC (in orange) vs. actual eigenmodes (in blue).}
\label{fig:modes}
\end{figure}
\section{Example: the torque-limited double pendulum}
\label{sec:double-pendulum}
In order to validate the EigenMPC approach, we apply it to the double pendulum with point masses at the tip of each link, whose inertia matrix $M(x)$ in local coordinates $x=\begin{bmatrix} \theta_1 &\theta_2 \end{bmatrix}^\top$ is    
\begin{equation}
    \begin{bmatrix}
    m_1l_1^2+m_2(l_1+l_2)^2-2m_2l_1l_2\bigl(1-\cos(\theta_2)\bigr) & * \\
    m_2\bigl(l_2^2+l_1l_2\cos(\theta_2)\bigr) & m_2l_2^2
    \end{bmatrix} \, ,
\end{equation}
and its gravitational potential is $V(x)=V^*(x)-V^*(0)$, where
\begin{equation}
    V^*(x)= -m_1gl_1\cos(\theta_1)-m_2g\bigl(l_1\cos(\theta_1)+l_2\cos(\theta_1+\theta_2)\bigr) \, ,
\end{equation}
with $(x,\dot{x})=0$ being a stable equilibrium configuration where the pendulum is stretched downwards. In addition, for the purpose of this work, its joint torques are limited to $\pm \SI{1}{Nm}$. 

As mentioned in \cite{albu-schaeffer2020review}, the pendulum has two families of eigenmodes evolving from the eigenvectors: the in-phase mode and the anti-phase one. In the next section, both eigenvectors will be used as \textit{a priori} knowledge to find their respective family of eigenmodes. 
\subsection{Simulation setup}
For the evaluation of the EigenMPC framework, a pendulum with both masses equal to $\SI{1}{kg}$ and both link lengths equal to $\SI{1}{m}$ was used. Furthermore, MATMPC \cite{chen2019matmpc} was applied to solve the NMPC program. 

% A dense SQP approximation of the NLP was set up and solved by qpOASES using the RTI scheme. The algorithm was executed on a OpenSUSE-based Linux computer equipped with Intel 3.70GHz Xeon E5-1620 v2 CPU (x8) and 8GB RAM. 

The EigenMPC gains were tuned once for the best performance and used in all simulations. The chosen gains were $w_x=50$, $w_{\dot{x}}=2500$ and $w_F=225$ for both versions, $w_E=25$ for the curved version, and $w_E=5$ for the straight version with $ \alpha=\beta=0.1$. The shooting interval for all simulations was $\SI{25}{ms}$ and the prediction horizon was $N=80$ shooting steps. \textcolor{black}{Among all experiments, the worst average computation time was $\SI{5.3}{ms}$ when run on a computer equipped with an Intel 3.70GHz Xeon E5-1620 v2 CPU (x8) and 8GB RAM.}

The numerical evaluation of the proposed approach is divided into two stages. The first aims at comparing the final trajectories found by the straight version with the eigenmodes computed by the framework presented in \cite{albu-schaeffer2020review}. The second analyzes the convergence capabilities and particularities of both versions of EigenMPC. A supplementary video is available at \url{https://doi.org/10.4121/19196765}.

\begin{figure}[t]
\centering
\begin{subfigure}[t]{0.44\linewidth}
    \centering
    \includegraphics[width=0.9\linewidth]{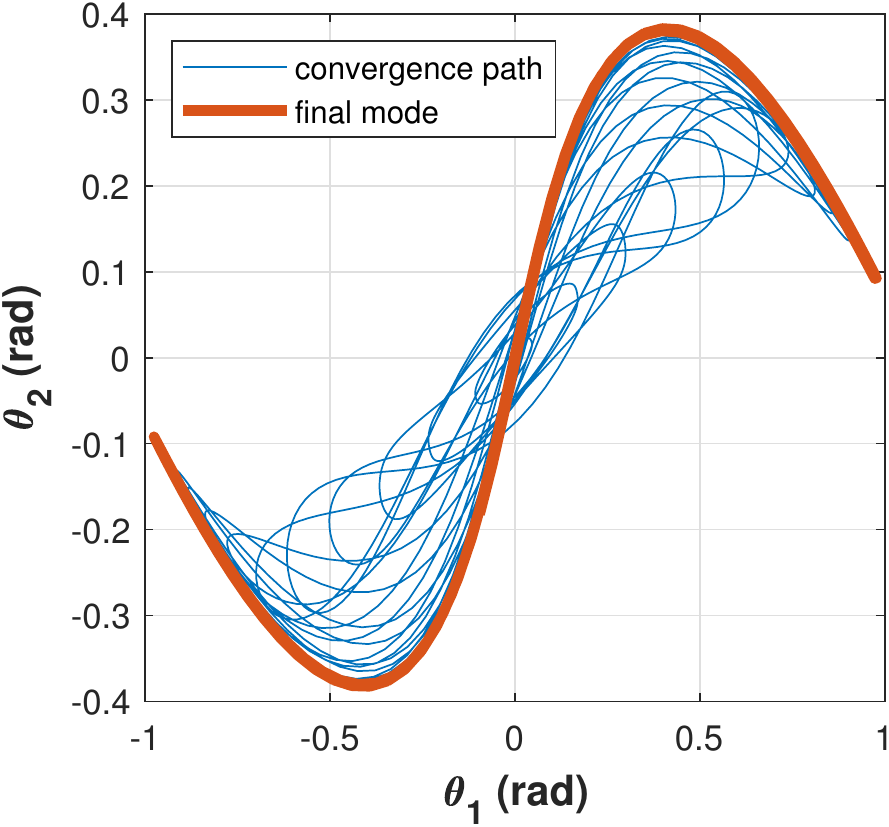}
    \caption{Joint-space path.} %{Light Unit}
    \label{subfig:straight_constrained_a}
\end{subfigure}%
\begin{subfigure}[t]{0.44\linewidth}
    \centering
    \includegraphics[width=0.9\textwidth]{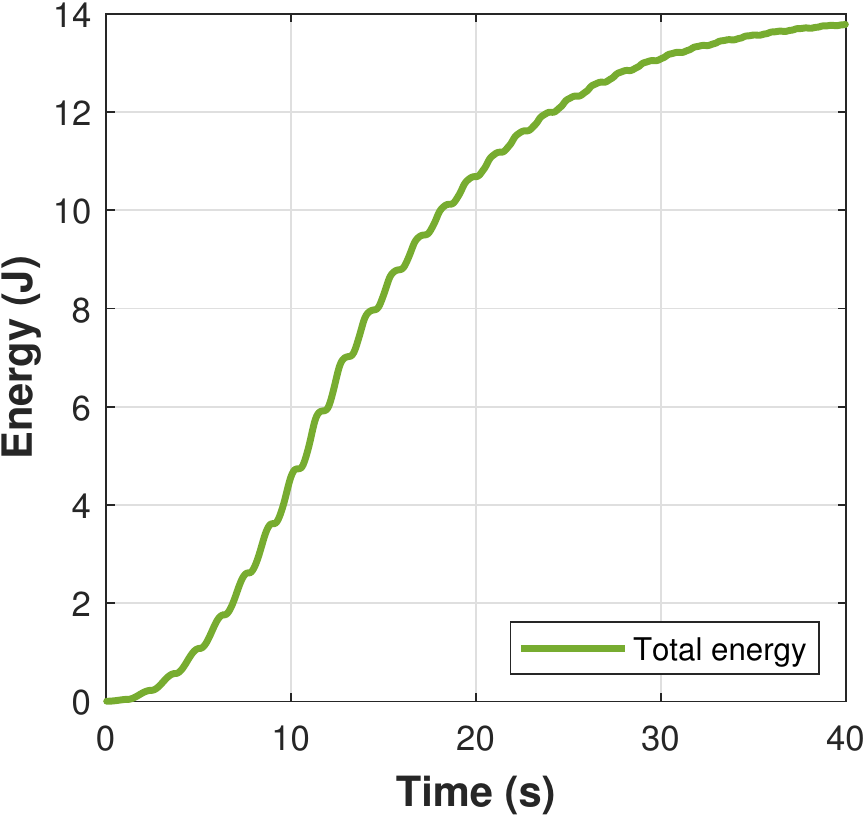}
    \caption{Total energy.}
    \label{subfig:straight_constrained_b}
\end{subfigure}

\begin{subfigure}[t]{0.44\linewidth}
\centering
 \includegraphics[width=0.9\textwidth]{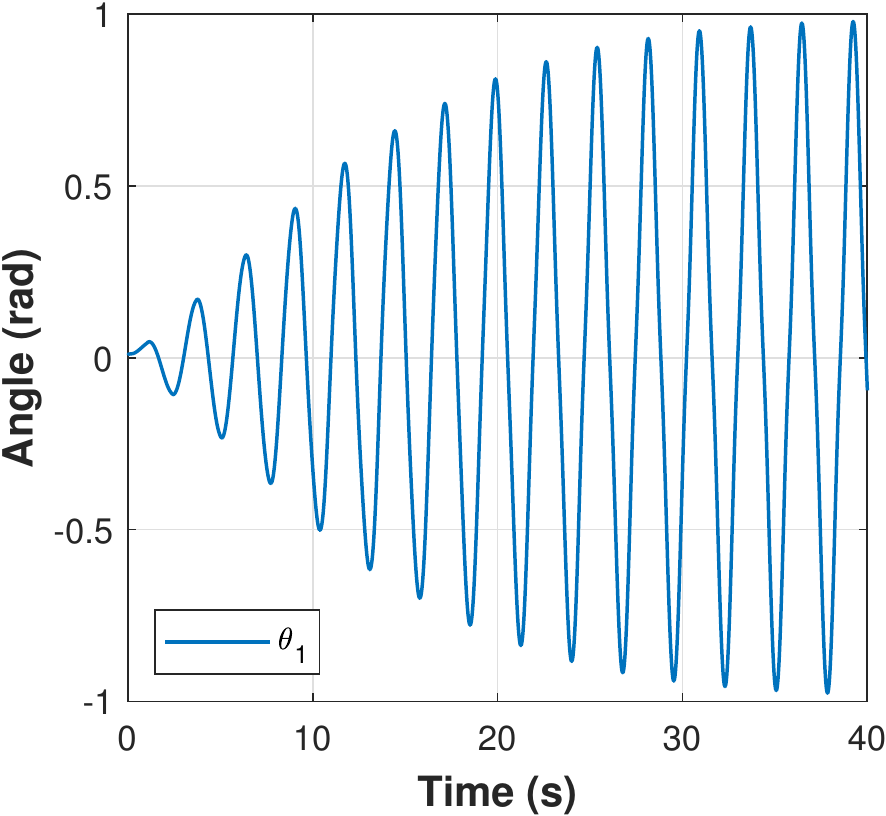}
\caption{$\theta_1$.}
\label{subfig:straight_constrained_c}
\end{subfigure}%
\begin{subfigure}[t]{0.44\linewidth}
\centering
 \includegraphics[width=0.9\textwidth]{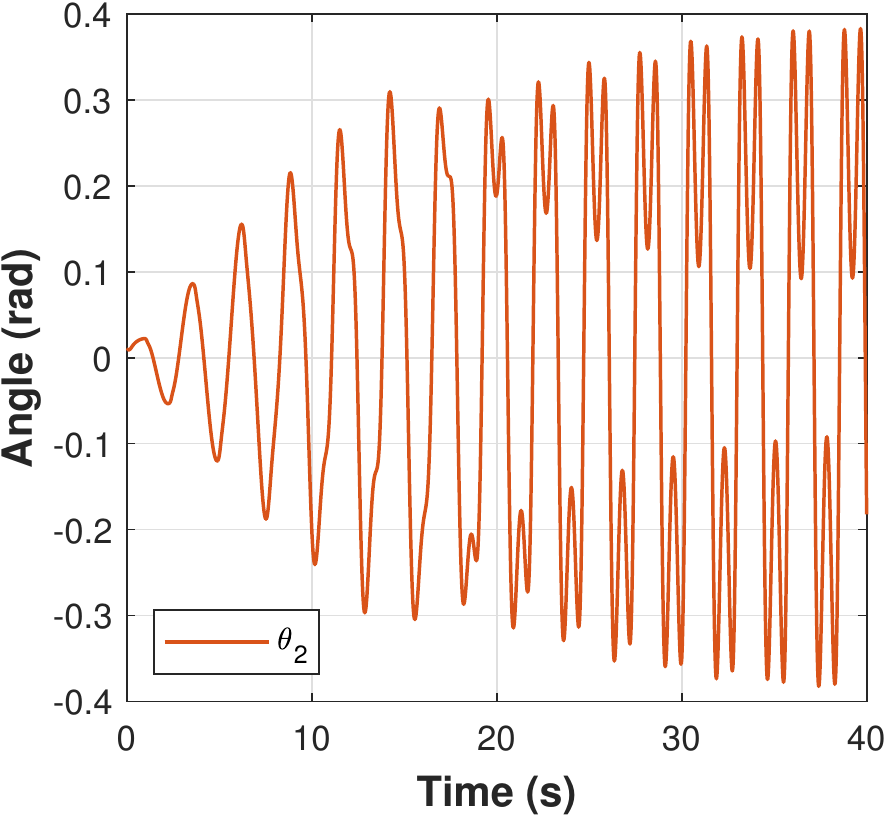}
\caption{$\theta_2$.}
\label{subfig:straight_constrained_d}
\end{subfigure}\\
\begin{subfigure}[t]{0.44\linewidth}
\centering
 \includegraphics[width=0.9\textwidth]{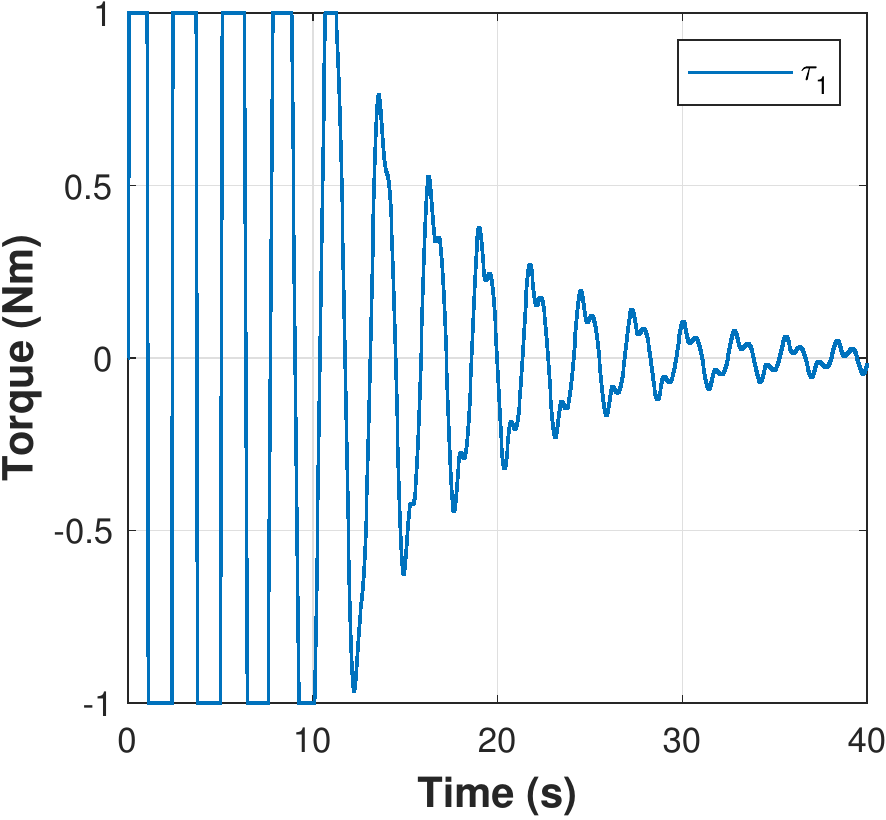}
\caption{$\tau_1$.}
\label{subfig:straight_constrained_e}
\end{subfigure}%
\begin{subfigure}[t]{0.44\linewidth}
\centering
 \includegraphics[width=0.9\textwidth]{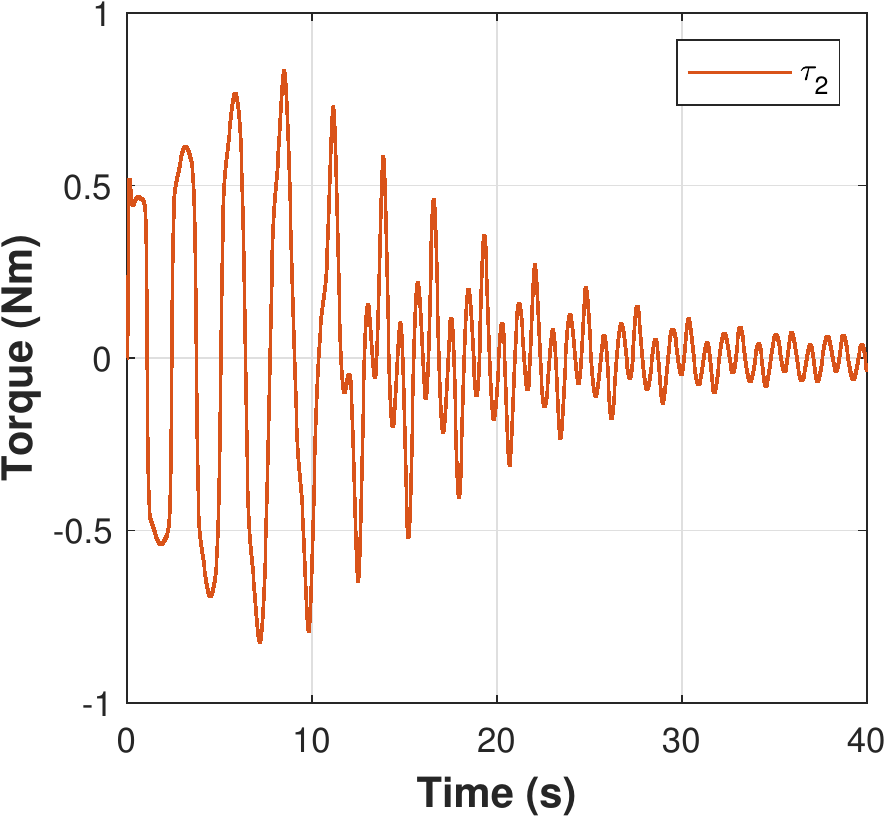}
\caption{$\tau_2$.}
\label{subfig:straight_constrained_f}
\end{subfigure}
\caption{Straight EigenMPC, $E_{ref} = \SI{14}{J}$, saturated controls.}
\label{fig:straight_constrained}
\end{figure}%

\begin{figure}[tb]
\centering
\begin{subfigure}[t]{0.44\linewidth}
    \centering
    \includegraphics[width=0.9\linewidth]{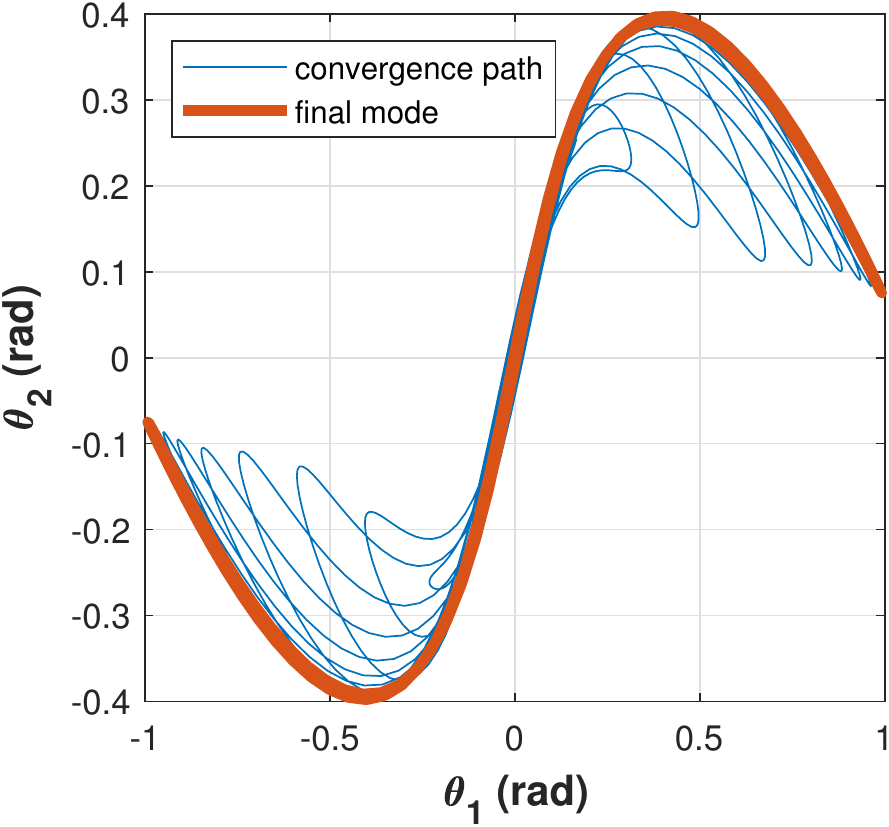}
    \caption{Joint-space path.} %{Light Unit}
    \label{subfig:curved_constrained_a}
\end{subfigure}%
\begin{subfigure}[t]{0.44\linewidth}
    \centering
    \includegraphics[width=0.9\textwidth]{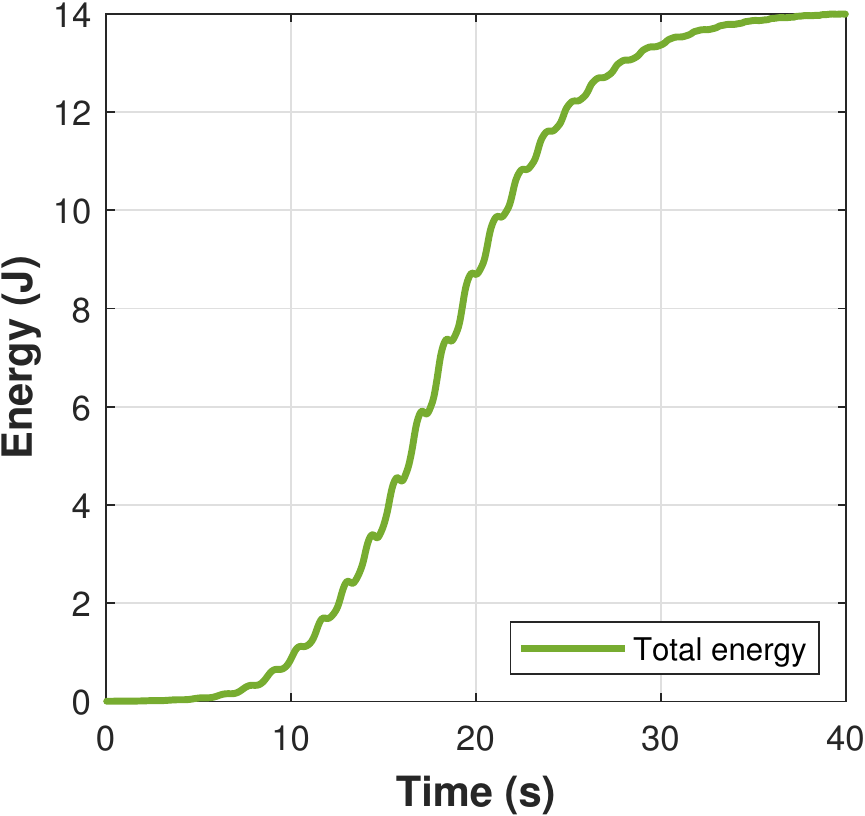}
    \caption{Total energy.}
    \label{subfig:curved_constrained_b}
\end{subfigure}

\begin{subfigure}[t]{0.44\linewidth}
\centering
 \includegraphics[width=0.9\textwidth]{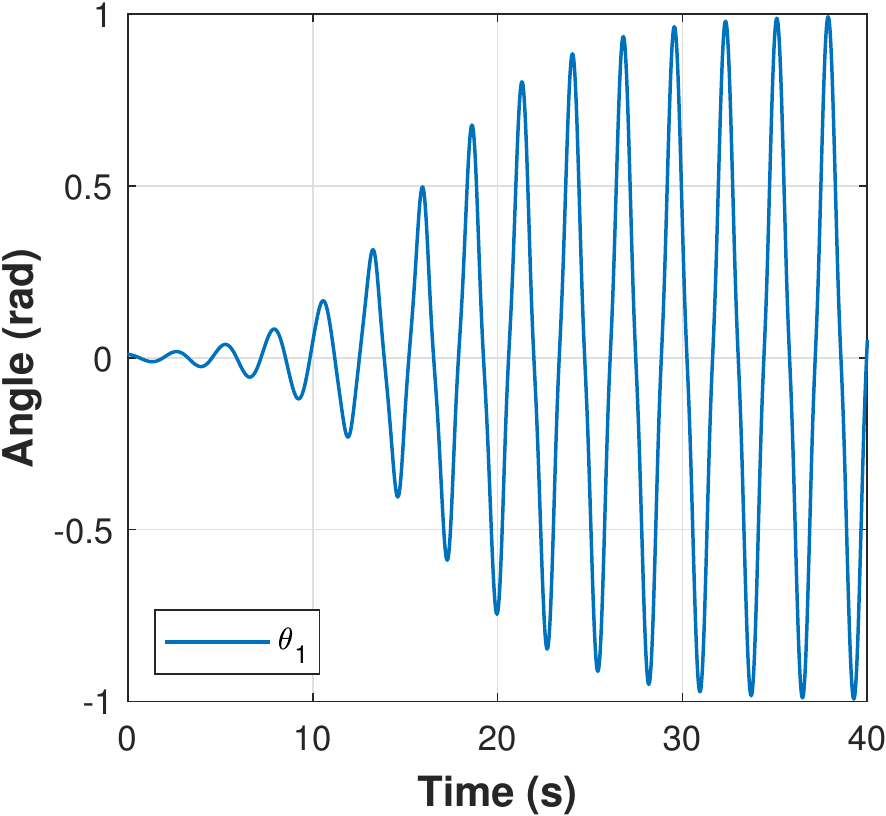}
\caption{$\theta_1$.}
\label{subfig:curved_constrained_c}
\end{subfigure}%
\begin{subfigure}[t]{0.44\linewidth}
\centering
 \includegraphics[width=0.9\textwidth]{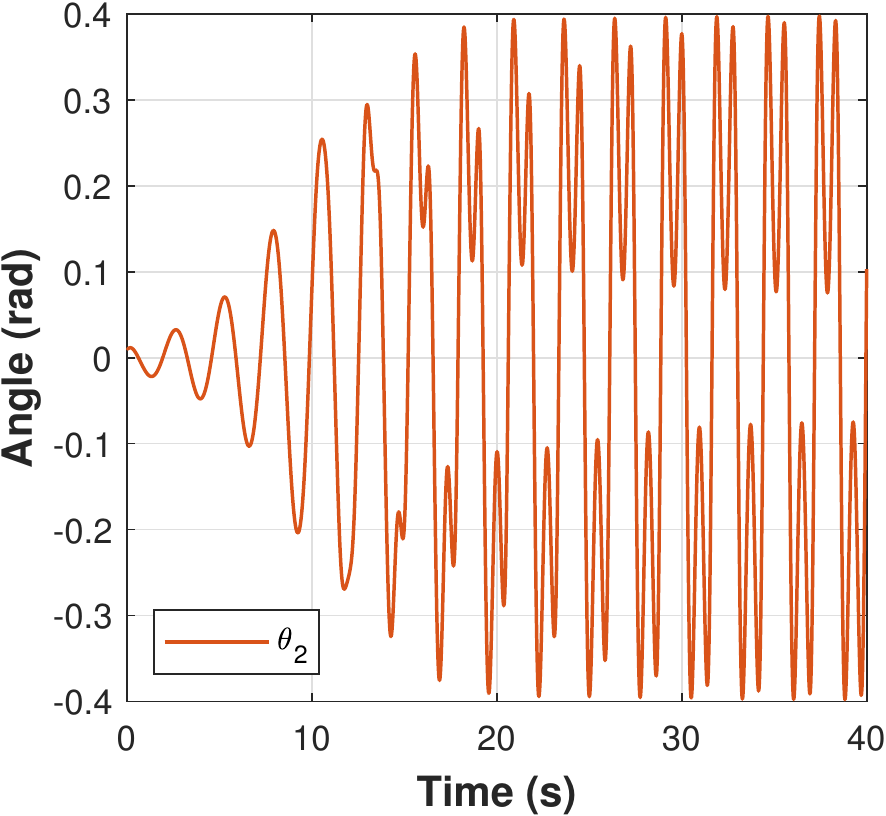}
\caption{$\theta_2$.}
\label{subfig:curved_constrained_d}
\end{subfigure}\\
\begin{subfigure}[t]{0.44\linewidth}
\centering
 \includegraphics[width=0.9\textwidth]{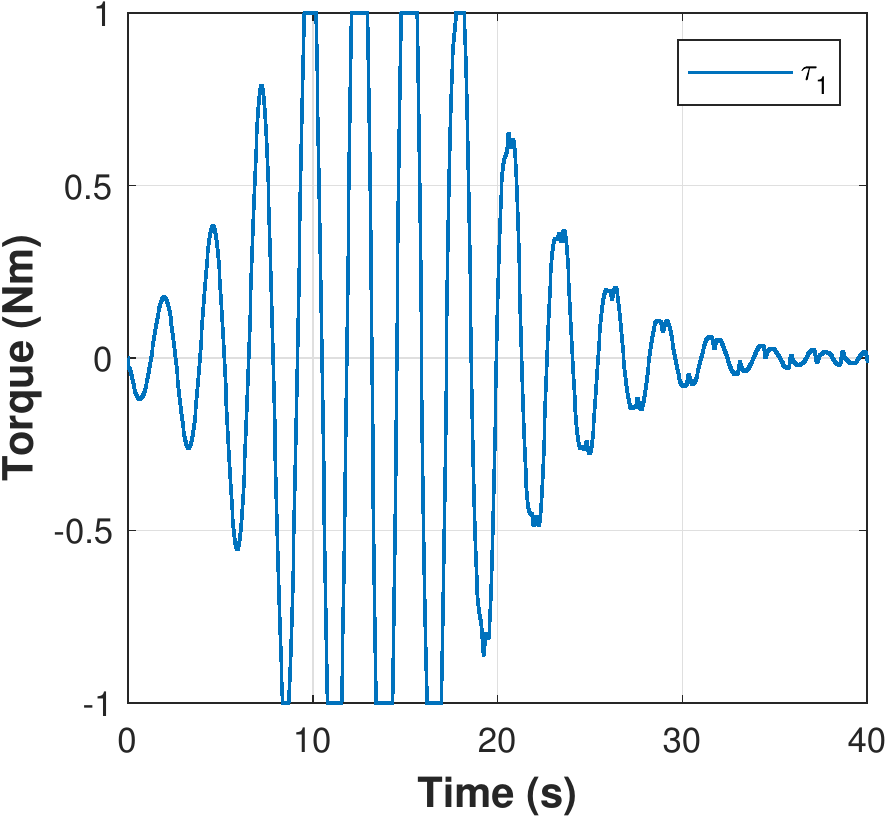}
\caption{$\tau_1$.}
\label{subfig:curved_constrained_e}
\end{subfigure}%
\begin{subfigure}[t]{0.44\linewidth}
\centering
 \includegraphics[width=0.9\textwidth]{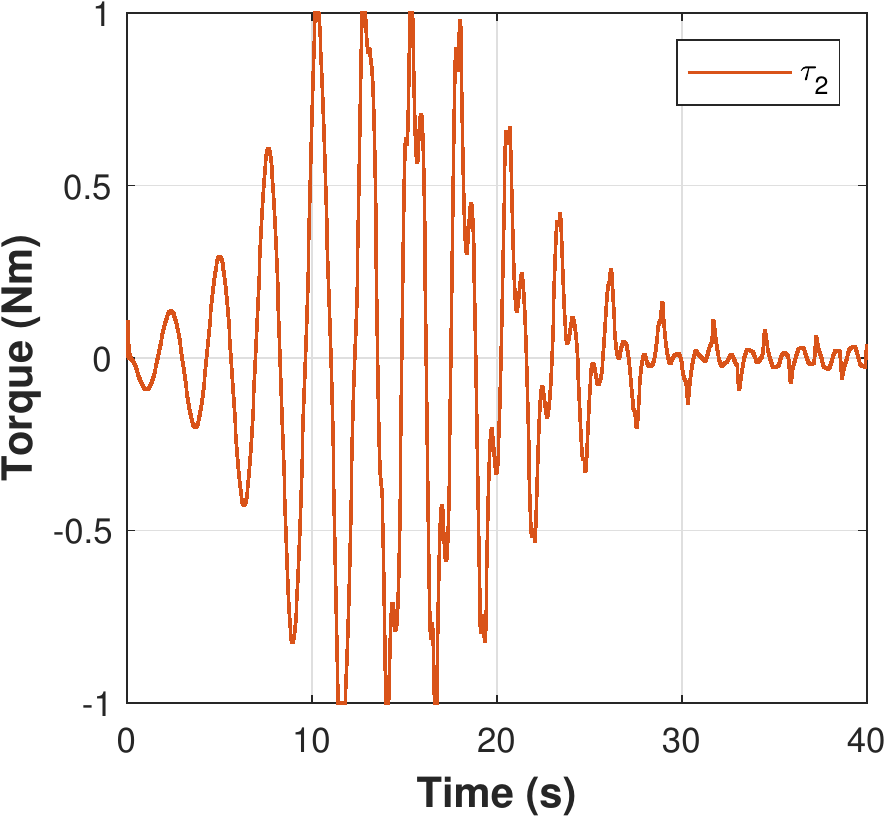}
\caption{$\tau_2$.}
\label{subfig:curved_constrained_f}
\end{subfigure}
\caption{Curved EigenMPC, $E_{ref} = \SI{14}{J}$, saturated controls.}
\label{fig:curved_constrained}
\end{figure}

\subsection{Eigenmode-finding capabilities}

In order to assess the similarity between the trajectories the straight version of EigenMPC converges to and the actual modes, for each of the modes (in-phase and anti-phase) two energy levels were given as reference. The first was $E_{ref}\approx\SI{2}{J}$, the second was $E_{ref} \approx \SI{16}{J}$ for the in-phase mode and $E_{ref} \approx \SI{12}{J}$ for the anti-phase one, which was the highest value for which a line-shaped anti-phase mode could be found by the straight version. 

The results are depicted in Fig.~\ref{fig:modes}. The proposed approach performs well in finding both more straight modes, like the anti-phase ones, and more curved ones, most notably the one in Fig.~\ref{subfig:modes_b}.
The most apparent mismatch is in the curvature of the path shown in Fig.~\ref{subfig:modes_a}, which is slightly more straight than the actual eigenmode. This happens because, for lower energies EigenMPC tends to converge to paths that look more like the original eigenvector.

Despite not having a formal asymptotic stability proof and being able to converge to a neighborhood of the eigenmodes only, the straight EigenMPC produces final results similar to those from the algorithm presented in \cite{albu-schaeffer2020review}. Therefore, it serves as an alternative eigenmode-search routine in practical applications.

\subsection{Eigenmode stabilization for constrained systems}
After showing that the proposed framework is capable of finding an approximation of the actual eigenmodes, this section aims at assessing its convergence properties in two different scenarios.
\subsubsection{Nominal case}
Initially, both versions of EigenMPC were applied in order to reach an in-phase eigenmode with $E_{ref}=\SI{14}{J}$, starting from the neighborhood of the equilibrium. While for the straight case, the in-phase eigenvector of the linearized system was used, for the curved one, a ninth-order polynomial was used to define $X$, $\dot{X}$ and $c$ based on the eigenmode found using \cite{albu-schaeffer2020review}.

The results are shown in Figs.~\ref{fig:straight_constrained} and \ref{fig:curved_constrained}. Figs.~\ref{subfig:straight_constrained_a} and \ref{subfig:curved_constrained_a} show the joint-space path ($\theta_1 \times \theta_2$) taken by the system until reaching its final trajectory, shown in orange. As expected, for the straight version, the system takes an increasingly curved path as the energy grows. On the other hand, the curved version takes a path whose curvature is similar to the eigenmode. Nevertheless, both versions reach similar final trajectories. It can also be noted that, although the final magnitude of the control action for both controllers is near zero, as expected, the curved version reaches even lower final torques since a chart to the actual eigenmode is used in the cost function. Moreover, due to the initially low cost on the control action, the straight version initially applies higher torques, which reduce as energy builds up. On the other hand, the curved version, having a constant cost on the control action, takes longer to leave the low-energy region. However, both controllers take about the same time to reach the desired energy.

\subsubsection{Large initial conditions}
\begin{figure}[t]
\centering
\begin{subfigure}[t]{0.46\linewidth}
\centering
 \includegraphics[width=0.9\textwidth]{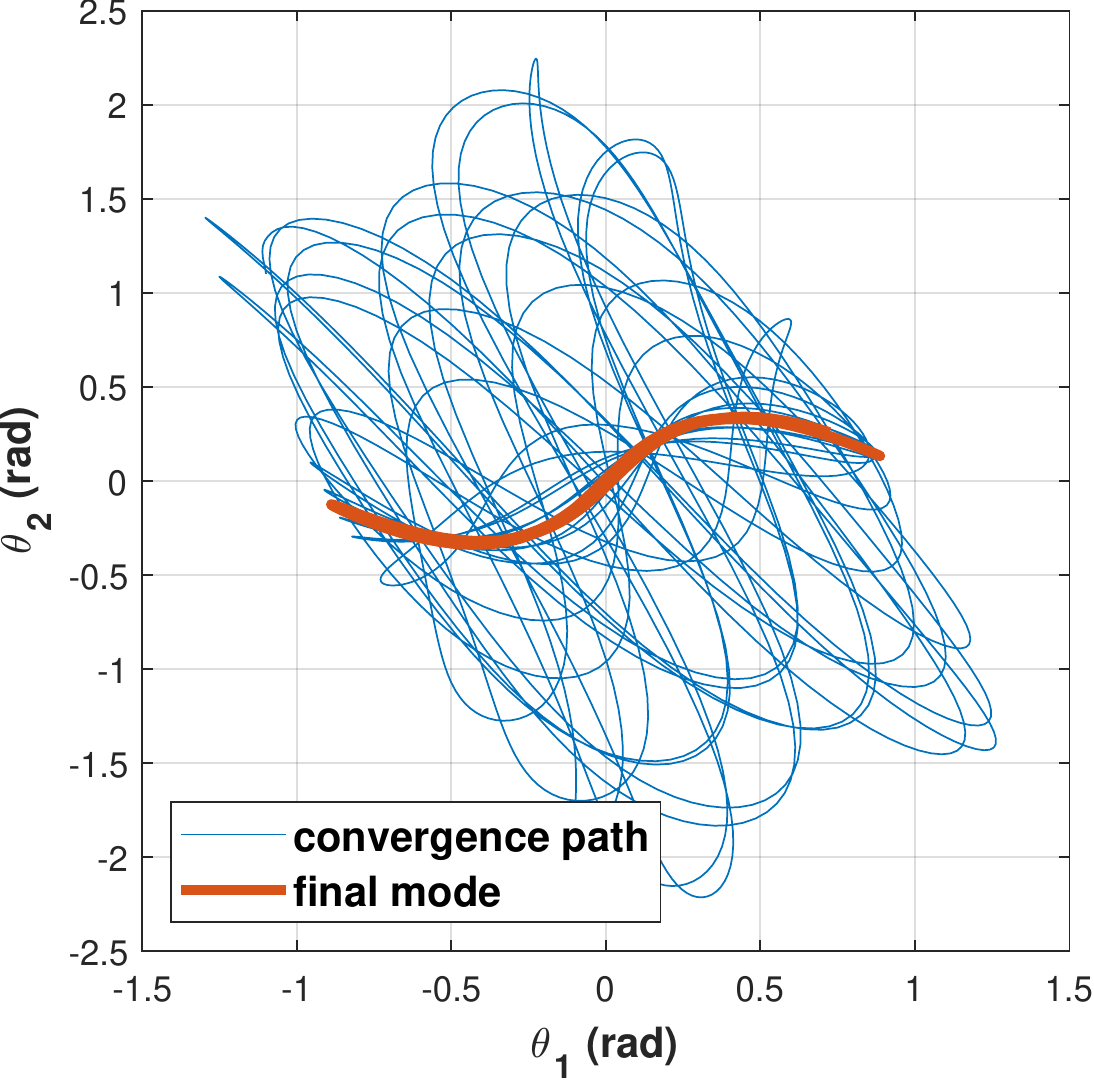}
\caption{Straight EigenMPC.}
\end{subfigure}%
\begin{subfigure}[t]{0.46\linewidth}
\centering
 \includegraphics[width=0.9\textwidth]{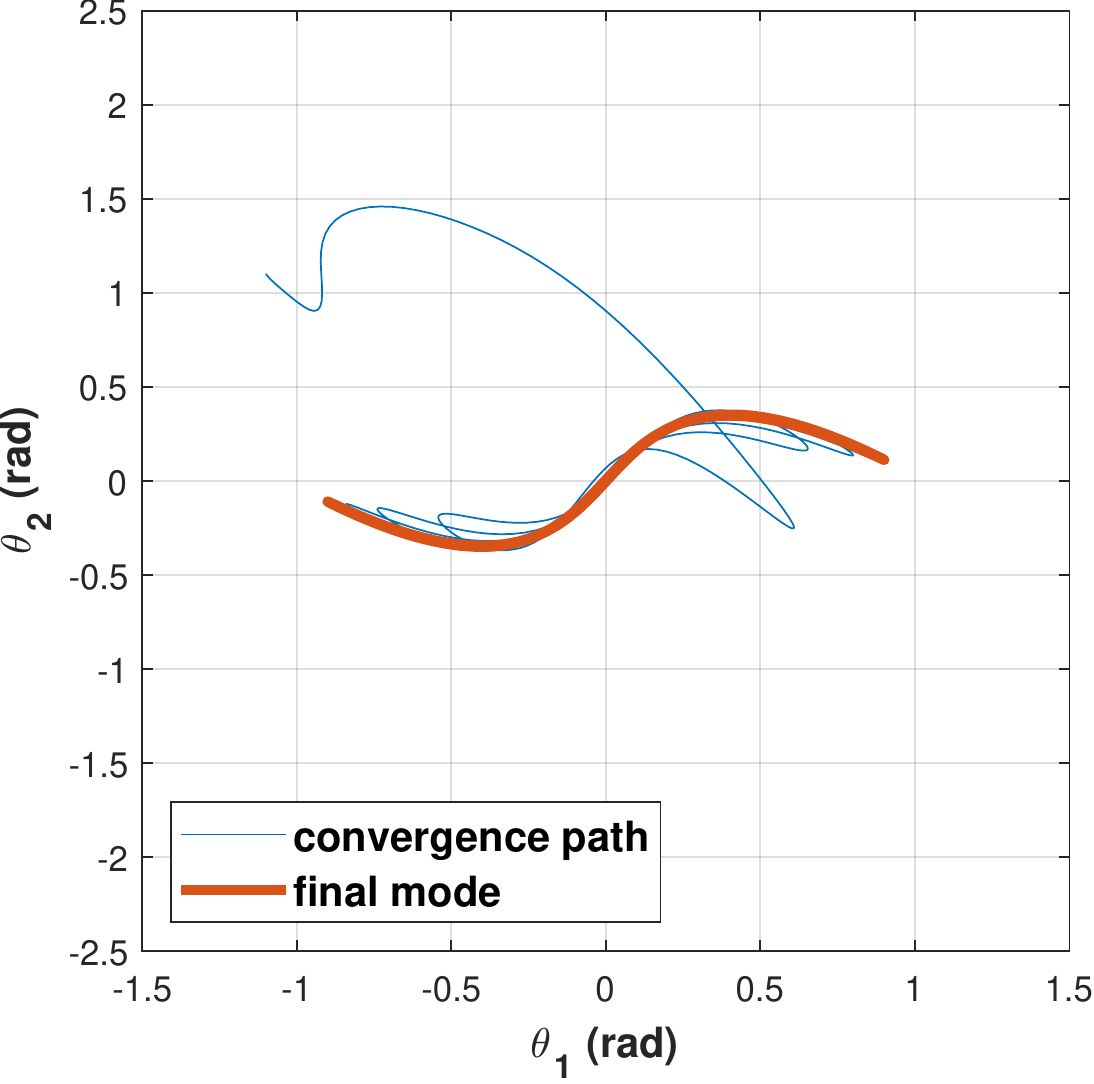}
\caption{Curved EigenMPC.}
\end{subfigure}
\caption{Convergence from large initial conditions.}
\label{fig:large_ic}
%\vspace{0.5cm}
%\begin{subfigure}[t]{0.46\linewidth}
%\centering
% \includegraphics[width=0.9\textwidth]{figures/path_energy_inphase0_curved_20percent.pdf}
%\caption{In-phase mode.}
%\end{subfigure}
%\begin{subfigure}[t]{0.46\linewidth}
%\centering
% \includegraphics[width=0.9\textwidth]{figures/path_energy_antiphase0_curved_20percent.pdf}
%\caption{Anti-phase mode.}
%\end{subfigure}
%\caption{Final path, eigenmode mismatch, hybrid EigenMPC.}
%\label{fig:20percent}
\end{figure}

Subsequently, aiming at assessing a possible weakness of the straight version, i.e., \textcolor{black}{low stiffness (weak position-error feedback) for high energies}, we simulated the system starting from larger initial conditions, namely $\theta=(-1.1,\,1.1) \text{ rad}$. The results are shown in Fig.~\ref{fig:large_ic}. As expected, due to its constantly high stiffness, the curved EigenMPC recovers quicker than its straight counter part. The straight version, on the other hand, due to its low stiffness at high energies, caused by the term $1-\tanh(\alpha E)$, takes longer to converge, but still reaches the final mode. 

%\subsubsection{Hybrid EigenMPC}
%In order to take advantage of the high stiffness of the curved controller, but still endowing it with some adaptability, a hybrid version of EigenMPC can also be used. In this case, the $\tanh$ terms can be added to \eqref{eq:eig_NLP_curved}. This is especially beneficial in case a computation of the eigenmode is available, but for a slightly different model (e.g., in case of parametric uncertainty or variation). To test that, $X$, $\dot{X}$, and $c$ used in the cost function of the controller were computed for a double pendulum with $m_2$ 20\% higher than in the actual system. In this case, $\alpha=0.01$ was used, which reduces the weight of the line-converging cost by only 10\% at $E=\SI{14}{J}$, but already gives the system some exploration characteristic. The final trajectories for both the in-phase and the anti-phase modes are shown in Fig.~\ref{fig:20percent}. As can be seen, the hybrid EigenMPC brought the system to the vicinity of the actual eigenmode in both cases, despite having used a different one in its cost function.   
\section{Conclusions and Future Work}
This paper presented a novel NMPC-based framework, which is capable of finding and converging to sustained nonlinear oscillations in the neighborhood of the eigenmodes or asymptotically converging to an eigenmode in case a pre-computed parametrization is available. Moreover, it guarantees energy efficiency not only in the final trajectory as \cite{della-santina21}, but also in the convergence phase, due to the suboptimality characteristic it inherits from the NMPC framework.
The limitation of the framework lies in the fact that no guarantee of global convergence to a desired mode can be given; however, it has been shown to recover well from significantly large initial conditions. \textcolor{black}{Future work will involve applying EigenMPC to more complex systems, like the DLR SAM, using more realistic propeller constraints, as in \cite{bicego20nonlinear}.}

\bibliographystyle{IEEEtran}
\bibliography{root}

\end{document}